\documentclass[letterpaper, 10 pt, conference]{ieeeconf}

\usepackage{subfigure}
\usepackage[utf8]{inputenc}
\usepackage[dvipdfmx]{graphicx}
\usepackage{amsmath}
\usepackage{url}
\usepackage{color}
\usepackage{cite}
\usepackage{algorithm}
\usepackage{algorithmic}
\usepackage{amssymb}
\usepackage{amsfonts}
\usepackage{multirow}
\usepackage{booktabs}

\IEEEoverridecommandlockouts
\title{\LARGE \bf
Extrinsic Calibration of Multiple LiDARs for a Mobile Robot\\ based on Floor Plane And Object Segmentation
}

\author{Shun Niijima$^{1}$, Atsushi Suzuki$^{1}$, Ryoichi Tsuzaki$^{1}$, Masaya Kinoshita$^{1} % 
$
\thanks{$^{1}$ Sony Group Corporation, Minato-ku, Tokyo, Japan, 108-0075 (email: shun.niijima@sony.com).
        {\tt\small }}%
}

\begin{document}

\maketitle
\thispagestyle{empty}
\pagestyle{empty}

%%%%%%%%%%%%%%%%%%%%%%%%%%%%%%%%%%%%%%%%%%%%%%%%%%%%%%%%%%%%%%%%%%%%%%%%%%%%%%%%
\begin{abstract}
  The utilization of mobile robots equipped with multiple light detection and ranging (LiDAR) sensors, capable of perceiving their surroundings, is on the rise due to the miniaturization and cost reduction of LiDAR technology. This paper introduces a target-less extrinsic calibration method for multiple LiDARs with non-overlapping fields of view (FoV). The proposed method leverages accumulated point clouds of the floor plane and objects obtained during robot motion. It enables accurate calibration, even in challenging configurations where LiDARs are directed towards the floor plane, which can introduce biased feature values. Additionally, the method incorporates a noise removal module that takes into account the scanning pattern to address bleeding points, which are significant sources of error in point cloud alignment when using high-density LiDARs. Evaluations conducted through simulation demonstrate that the proposed method achieves higher accuracy in extrinsic calibration with two and four LiDARs compared to conventional methods, regardless of the type of objects. Furthermore, experiments conducted using a real mobile robot validate the effectiveness of our proposed noise removal module in precisely eliminating noise compared to conventional methods. The estimated extrinsic parameters successfully contribute to the creation of consistent 3D maps.
\end{abstract}
%%%%%%%%%%%%%%%%%%%%%%%%%%%%%%%%%%%%%%%%%%%%%%%%%%%%%%%%%%%%%%%%%%%%%%%%%%%%%%%%

\section{Introduction}
\begin{figure}[!t]
\centering
\includegraphics[width=0.65\columnwidth, angle=0]{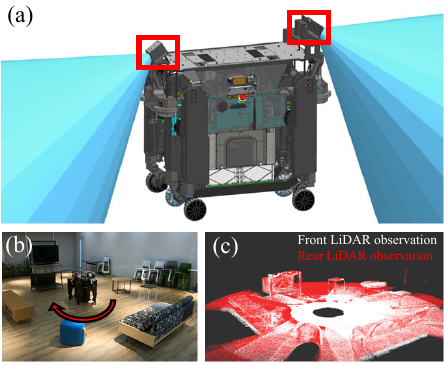}
\caption{
The proposed method achieves extrinsic calibration of multiple LiDARs with non-overlapping FoV such as mobile robot in (a). Our approach is accumulating data while moving similar to (b) and creating a common FoV to evaluate the consistency of the 3D map. The proposed method  achieves accurate extrinsic calibration and generates a consistent 3D map even when most of the observations are on the floor, shown in (c).
}
\label{fig:proposed_image}
\end{figure}

A light detection and ranging (LiDAR) is recognized as a critical sensor in autonomous driving technology due to its high accuracy observation, and it has been advancing in miniaturization and cost reduction.  Consequently, the incorporation of LiDARs into mobile robots is increasing.  The spreading of high-density LiDARs \cite{liu2021low} is also enhancing the perception capabilities of mobile robots.  However, it is difficult for a single LiDAR to cover the entire surroundings of a mobile robot, including the travel surface. It is common to mount multiple LiDARs in different directions, such as front and back or left and right, as shown in Fig. \ref{fig:proposed_image}(a). In the future, it is expected that widespread adoption of these mobile robots equipped with multiple LiDARs. With the proliferation of such mobile robots, there arises a need for robot inspection sites around the world to calibration at the time of shipment, time of deterioration over time, after overhauls and so on.

Accurate integration of data from multiple LiDARs requires robust extrinsic calibration.  A general and useful method involves observing a common object with distinctive features such as spheres or planes simultaneously with multiple LiDARs \cite{xue2019automatic}\cite{9812062}.  However, since each LiDAR is positioned to recognize different directions, their field of view (FoV) rarely overlap, if at all.  Therefore, placing a common object in the shared FoV is difficult for actual mobile robots, making these target-based methods challenging to apply. 

To avoid additional tasks such as limitations of the Field of View (FoV) and the preparation of specific components, methods have been proposed that create a common FoV among sensors by accumulating point clouds while in motion \cite{liu2021extrinsic}\cite{liu2022targetless} similar to Fig. \ref{fig:proposed_image}(b).  These methods minimize observation errors between LiDARs by using the entire set of environmental features such as planes and line features accumulated during movement.  However, for robots that travel on the ground, LiDARs are often mounted to detect the floor surface, resulting in observations that are predominantly of the floor shown in Fig. \ref{fig:proposed_image}(c).  Consequently, the features obtained may be insufficient, or the floor features may become dominant, leading to unsuccessful extrinsic calibration.

This paper proposes a method  estimating the extrinsic parameters of multiple LiDARs with non-overlapping FoV by utilizing point clouds of floor plane and objects accumulated while in motion. Our method pre-segment the floor plane and object point clouds, and perform a two-step estimation process that utilizes each point cloud.  It achieves accurate extrinsic calibration even when the majority of LiDAR observations are of the floor.  

Furthermore, as reported in this research \cite{yuan2021pixel}, high-density LiDARs are known to exhibit prominent bleeding points, which are generated as they extend from the edge of the objects.  This noise is caused by the mixing of pulses hitting both foreground and background objects. 
These noises are a major source of error in extrinsic calibration. Our method includes a module that can remove noise more accurately compared to conventional methods, based on a very simple concept.

Our contributions are as follows:
\begin{itemize}
\item We propose the extrinsic calibration method for multiple LiDARs with non-overlapping FoV.  Our method estimates the movement trajectory and extrinsic parameters simultaneously using segmented floor and object point clouds. It enables accurate calibration even when floor information is dominant and challenging for conventional methods.
\item We propose a simple noise removal module useful for calibrating with high-density LiDARs.
\item Through evaluation in simulation of differents environments where various types of objects are placed on the floor plane, and on differents configurations of LiDARs with non-overlapping FoV,  we demonstrate that our proposed method achieves higher accuracy than the conventional methods.
\item The experiments using a real mobile robot has shown that our proposed noise removal module can eliminate noise more precisely than conventional methods, and the estimated extrinsic parameters have successfully created consistent 3D maps by LiDARs with non-overlapping FoV.
\end{itemize}

\section{Related work}

\subsubsection{Multiple LiDAR calibration}
There are two main approaches to multiple LiDAR calibration: target-based and targetless methods.  Target-based methods estimate extrinsic parameters by preparing special components as targets to assist in feature extraction and observing them simultaneously with multiple LiDARs. These research use retro-reflective targets \cite{xue2019automatic}, specially processed plates \cite{domhof2019extrinsic}, or spherical objects \cite{9812062} to aid in the extraction of 3D features such as edges, planes, and cones. While target-based methods have relatively low computational complexity, they require the creation of specific targets, which necessitates additional time for preparing.  Since the mobile robot may frequently perform extrinsic calibration in various locations, it is desirable to avoid the need to prepare special components.
In targetless calibration methods, there are two representative approaches: one is motion-based that estimates extrinsic parameters from motion, and the other is feature-based extracts environmental features to estimate extrinsic parameters. Motion-based approaches \cite{lin2020decentralized, heng2020automatic} assume that each sensor undergoes the same rigid body motion at each time interval, transforming the extrinsic calibration into a hand-eye problem. Methods have also been proposed that facilitate LiDAR motion estimation by introducing other external sensors such as GNSS or INS \cite{chang2023versatile}\cite{levinson2014unsupervised}. The calibration accuracy of these approaches is susceptible to the precision of the sensor's motion estimation results, which may compromise reliability.

Feature-based methods estimate extrinsic parameters without special components by extracting features within the environment.  In particular, studies \cite{fernandez2015extrinsic,jiao2019novel} detect planar features in natural environments to obtain correct associations. Research \cite{fernandez2015extrinsic} imposes coplanarity and perpendicularity constraints on line segments extracted from vertical planes encountered in structured scenes to constrain the 6 degrees of freedom (6DoF) parameter space. Study \cite{jiao2019novel} uses three linearly independent planes, which are easier to obtain than vertical planes. These methods are bound by the same limitation as target-based methods: a common FoV is essential. 

To solve these problems, methods have been proposed that collect environmental features while in motion to create a common FoV among LiDARs with no FoV overlap \cite{liu2021extrinsic}. 
These methods perform global optimization of a graph composed of extrinsic parameters of each LiDAR and movement trajectories to ensure the consistency of the collected point cloud data.
 Studies \cite{liu2022targetless}\cite{jiao2019automatic} have been modified to extract plane and line segment features by adaptive size voxels to improve computational efficiency,
   and achieves an average translational error of 0.01m and a rotational error of less than 0.1 degrees.

However, these methods needs environments that have sufficient and various direction of features to constrain the 6DoF, such as wall and ceiling.  When applying these methods to ground-traveling robots as shown in Fig. \ref{fig:proposed_image}(a), the point cloud data obtained consists only of the ground and a few objects, leading to insufficient features or dominance of floor point clouds, which can result in unsuccessful extrinsic calibration.  To address these issues, our method pre-segment the floor plane and object point clouds, and adopt a two-step estimation process that utilizes each point cloud, 
  and achieves comparable accuracy to above existing methods.

\subsubsection{LiDAR noise reduction}
These are two main approaches to noise reduction in LiDAR: point cloud-based and image-based.  In point cloud-based methods, among the effective point cloud noise removal methods are the statistical outlier removal filter (SOR) \cite{SOR} and the radius outlier removal filter (ROR) \cite{ROR}. These filters remove noise by identifying areas with low surrounding point density. This is based on the characteristic that noise occurs randomly, and its density is sparser than that of points that have hit an object.  The aforementioned methods are most effective when the density of the point cloud is consistent regardless of location. However, the density of point clouds observed by LiDAR becomes sparser with increasing distance from the sensor, which means that directly applying these methods to LiDAR data may not yield adequate performance.

To address these issues, the dynamic statistics outlier filter (DSOR) \cite{DSOR} and the dynamic radius outlier removal filter (DROR) \cite{DROR} have been proposed. These methods dynamically vary the threshold for noise determination according to the distance from the sensor, considering that point density is higher closer to the sensor and lower farther away, successfully removing even nearby dense noise.  High-density LiDARs use scanning patterns that are not constant with each scan, so the density of point clouds is not simply dependent on the distance from the sensor but is also significantly affected by the sensor pattern. For example, LiDARs from Livox use a non-repetitive scanning method \cite{liu2021low}.

\begin{figure*}[!t]
\centering
\includegraphics[width=1.7\columnwidth, angle=0]{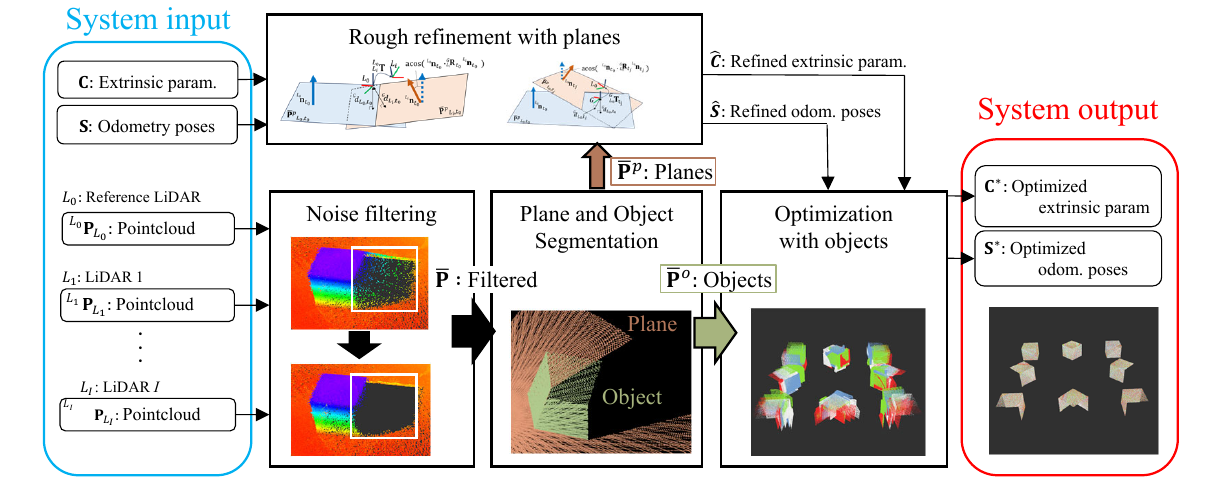}
\caption {System overview of proposed multiple LiDAR calibration method}
\label{fig:system_overview}
\end{figure*}

Another approach involves converting the observed point clouds from LiDAR into depth image format and applying computer vision or deep learning for noise classification from LiDAR.  High-density LiDARs by non-repetitive scanning \cite{liu2021low} obtain dense observation point clouds through the overlay of sparse point clouds, making it difficult to create dense depth images with all values filled, which can hinder the application of the aforementioned image-based methods.  These images are fed into the training model to detect features to identify noise from the images. Whethernet\cite{heinzler2020cnn} and 4DenoiseNet\cite{seppanen20224denoisenet} have been shown to be effective in improving data quality from LiDAR with this approach.  In certain scenarios, deep learning methods may outperform traditional approaches. However, their performance heavily depends on large-scale labeled datasets, which can be ineffective when labeled data is limited. 

In this paper, we propose a novel point cloud-based noise filter that dynamically changes the noise threshold considering the scanning pattern to efficiently remove noise from LiDARs that obtain non-constant observations with each scan.

\section{Methods}
\subsection{Problem setting}
The goal of extrinsic calibration for multiple LiDARs is to determine the coordinate transformations ${\rm \bf C} = \{^{L_0}_{L_1}{\rm \bf T}, \cdots, ^{L_0}_{L_{I}}{\rm \bf T} \}$ between a reference LiDAR $L_0$ and other LiDARs $\{L_1 \cdots L_{I}\}$.
Here, $^{A}_{B}{\rm \bf T} = (^{A}_{B}{\rm \bf R}, ^{A}_{B}{\rm \bf t}) \in SE(3)$ represents the coordinate transformation from frame A to frame B, consisting of a rotation component $^{A}_{B}{\rm \bf R} \in SO(3)$ and a translation component $^{A}_{B}{\rm \bf t} \in \mathbb{R}^3$.

To create a common FoV among LiDARs with non-overlapping FoV, the mobile robot accumulates data at multiple timestamps by alternating between stopping and moving. For formalization, assume the robot captures movement trajectory ${\rm \bf S} = \{^{G}_{L_0}{\rm \bf T}_{t_0} \cdots ^{G}_{L_0}{\rm \bf T}_{t_{J}} \}$ denotes the global frame of reference for the movement trajectory.

To assess the consistency of the 3D maps constructed from point clouds accumulated by the reference LiDAR $\{L_0\}$ and other LiDARs $\{L_1 \cdots L_{I}\}$ while in motion, these point clouds are transformed into a single global coordinate system.
The global coordinates of LiDAR $L_i$ at each timestamp $t_j$ can be expressed using the movement trajectory $^{G}_{L_i}{\rm \bf T}_{t_j}$ and the coordinate transformation $^{L_0}_{L_i}{\rm \bf T}$ as follows:

\begin{equation}
^{G}_{L_i}{\rm \bf T}_{t_j} = ^{G}_{L_0}{\rm \bf T}_{t_j} \  ^{L_0}_{L_i}{\rm \bf T}
\label{eq:local_to_global_transformation}
\end{equation}

 The point cloud observed by LiDAR $L_i$ at time $\rm t_j$ in the LiDAR $L_i$ coordinate system is denoted as $^{L_i}{\rm \bf P}_{L_i, t_j}$, and the corresponding point cloud in the global coordinate system $^{G}{\rm \bf P}_{L_i, t_j}$ is obtained as follows:

\begin{equation}
\begin{split}
^{G}{\rm \bf P}_{L_i, t_j} &= ^{G}_{L_i}{\rm \bf T}_{t_j} \ ^{L_i}{\rm \bf P}_{L_i, t_j} \\
& = ^{G}_{L_i}{\rm \bf R}_{t_j} \ ^{L_i}{\rm \bf p}_{L_i,t_j} + ^{G}_{L_i}{\rm \bf t}_{t_j},
\end{split}
\label{eq:local_to_global_pointcloud}
\end{equation}
where ${\rm \bf p}_{L_i,t_j} = \{{\rm x,y,z}\}$ represents the coordinates of each point.

The problem of extrinsic calibration for multiple LiDARs with non FoV can be described as finding the optimal movement trajectory ${\rm \bf S}^*$ and coordinate transformations ${\rm \bf C}^*$ that minimize the error in the 3D maps constructed from each LiDAR's point cloud represented in the global coordinate system. 
Let $\rm e(^{G}{\rm \bf P}_{L_0, t},^{G}{\rm \bf P}_{L_i, t})$ be a function representing the error between the 3D map constructed by the reference LiDAR and the 3D map constructed by other LiDARs, the problem can be formulated as follows:

\begin{equation}
  ({\rm \bf C}^*, {\rm \bf S}^*) = \arg \mathop{\min}_{{\rm \bf C},{\rm \bf S}} \sum_{i=1}^{I}{\rm e(^{G}{\rm \bf P}_{L_0},^{G}{\rm \bf P}_{L_i})}
  \label{eq:graph_optimization}
\end{equation}
Various functions can be used to represent the error, such as point-to-point \cite{besl1992method}, point-to-plane \cite{chen1992object}, or quadratic entropy \cite{renyi1961measures}. Additionally, to improve computational efficiency, only 3D features may be used in advance \cite{liu2021extrinsic}.

\subsection{System overview}
LiDARs mounted to observe the travel surface predominantly capture point clouds of the floor. While the floor plane provides a strong constraint, it can lead to an optimization problem where only the alignment of the floor is prioritized, potentially failing to find the correct movement trajectory and extrinsic parameters. To address this issue, our process is divided into a rough refinement process using the floor point cloud and an optimization process using the objects point cloud with the floor removed. 

The system overview of the proposed method is illustrated in Fig. \ref{fig:system_overview}. Each module is represented by a rectangle. The inputs are the observed point clouds $^{L_i}{\rm \bf P}_{L_i}$ from the LiDARs, initial extrinsic parameters $\rm \bf C$, and the movement trajectory $\rm \bf S$. 
First, the noise removal module outputs denoised point cloud $^{L_{i}}{\rm \bf \bar{P}}_{L_{i}}$ from each LiDAR $^{L_{i}}{\rm \bf P}_{L_{i}}$. Next, the point clouds are divided into planes$^{L_{i}}{\rm \bf \bar{P}}^p_{L_{i}}$ and objects $^{L_{i}}{\rm \bf \bar{P}}^o_{L_{i}}$by the segmentation module. There are no constraints on the segmentation method, this paper uses simple RANSAC and Euclidean clustering. 
Subsequently, the rough refinement module uses the plane point clouds $^{L_{i}}{\rm \bf \bar{P}}^o_{L_{i}}$ to coarsely correct the movement trajectory $\rm \bf S$ and initial extrinsic parameters $\rm \bf C$, and outputs rough refine extrinsic parameters $\rm \bf \hat{C}$ and $\rm \bf \hat{S}$. 
Here, the movement trajectory and extrinsic parameters are corrected using the fact that LiDARs observing the travel surface are observing the same plane.
 Details are explained in Section III.D. 
Finally, the optimization module outputs the optimized set of extrinsic parameters $\rm \bf C^*$ and movement trajectory $\rm \bf S^*$ by solving equation (\ref{eq:graph_optimization}) using the object point clouds $^{L_{i}}{\rm \bf \bar{P}}^o_{L_{i}}$.  Details are discussed in Sections III.E.

\begin{figure}[!t]
\centering
\includegraphics[width=0.30\columnwidth, angle=0]{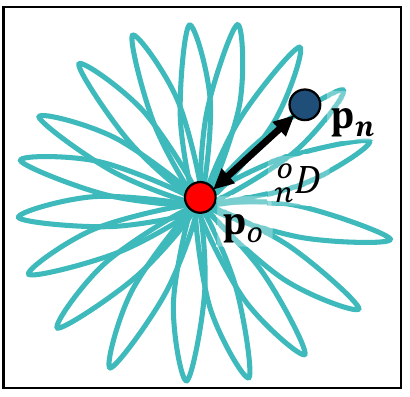}
\caption {Scanning pattern of a non-repetitive LiDAR}
\label{fig:livox}
\end{figure}

\begin{figure}[!t]
\begin{algorithm}[H] 
    \caption{Overview of the noise filter algorithm}
    \label{fig:NoiseFilterAlg}
    \begin{algorithmic}[1]    
    \renewcommand{\algorithmicrequire}{\textbf{Input:}}
    \renewcommand{\algorithmicensure}{\textbf{Output:}}
      \REQUIRE Point cloud: ${\mathbf P}=\{{\mathbf p}_0 \cdots {\mathbf p}_n\}, {\mathbf p}_n=(x_n,y_n,z_n)$
    \ENSURE  Filtered point cloud: ${\bar{\mathbf{P}}}$
    \STATE ${\mathbf P}$ $\leftarrow$ KdTree
    \FOR {${\mathbf p}_n$ $\in$ ${\mathbf{P}}$}
      \STATE mean distances $\leftarrow$ nearestKSearch(k)
    \ENDFOR
    \STATE calculate mean ($\mu$) and std. ($\sigma$) $\leftarrow$ mean distances
    \STATE calculate global threshold ($H_g$) $\leftarrow$ $\mu + (\sigma \times C_s)$
    \STATE determine observation center (${\mathbf p}_o$)
    \FOR {${\mathbf p}_n$ $\in$ ${\mathbf{P}}$}
        \STATE Dis. ($^o_n{D}$) $\leftarrow$ $\sqrt{(x_n - x_o)^2+(y_n - y_o)^2 + (z_n - z_o)^2}$
        \STATE calculate dynamic threshold ($H_d$) $\leftarrow$ $H_g \times C_r \times ^o_n{D}$
            \IF{mean distance $(\mu_n) < H_d$}
                \STATE ${\bar{\mathbf{P}}} \leftarrow {\mathbf{p}}_n$
            \ENDIF
    \ENDFOR
    \RETURN ${\bar{\mathbf{P}}}$
    \end{algorithmic}
\end{algorithm}
\end{figure}

\subsection{Noise filter for non-repetitive scan}
The purpose of the noise filter module is to output point clouds $^{L_i}{\rm \bf \bar{P}}_{L_i,t_j}$  from LiDAR observations $^{L_i}{\rm \bf P}_{L_i,t_j}$ that have been denoised. The proposed noise removal module follows the traditional filter concept that noise occurs randomly, and thus its density is sparser than that of point clouds from actual objects. DSOR considers the dynamic change in the density of observed point clouds with distance from the LiDAR, varying the noise threshold dynamically with distance. However, we focus on the fact that for LiDARs to observe the travel surface, where the distance between the sensor and observation points does not change significantly, the density of the LiDAR's point clouds is more dominantly affected by the LiDAR's scanning pattern than by the distance to the observation points. For example, Livox's non-repetitive scanning method, as shown in Fig. \ref{fig:livox}, results in varying densities depending on the distance from center of observation.

We propose a noise removal module that dynamically changes the noise threshold based on the distance  from the sensor's observation center approximating that the density of these scanning methods simply changes with the distance 
$^{o}_{n}{\rm D}$ between the observation center ${\rm \bf p}_o$ and the observation points ${\rm \bf p}_n$. The algorithm overview is shown in Algorithm \ref{fig:NoiseFilterAlg}.

\begin{itemize}
\item Preliminary preparation: First, a kd-tree is constructed for the observation point cloud (L.1). Then, for each point
  ${\rm \bf p}_n =\{x_n, y_n, z_n\}$,
the average distance to $k$ nearest neighbors is calculated (L.2-4). The average distance $\mu$ and standard deviation $\sigma$ are also calculated for the entire point cloud (L.5).
\item Global threshold calculation: Using the calculated average distance $\mu$ and standard deviation $\sigma$ for the entire point cloud, the global threshold $\rm {{H}}_g$ is calculated (L.6).
\begin{equation}
  \label{eq:Hg}
    {\rm H}_g= \mu+\sigma \times {\rm C}_s
\end{equation}
Here, ${\rm C}_s$ is the constant multiplication factor for standard deviation.
\item Noise estimation: Next, the observation center ${\rm \bf p}_o = \{x_o, y_o, z_o\}$ is determined (L.7). For each point ${\rm \bf p}_i$, the distance $^{o}_iD$ to the observation center ${\rm \bf p}_o$ is calculated (L.9).  
\begin{equation}
  \label{eq:D}
^{o}_nD =\sqrt{(x_n - x_o)^2+(y_n - y_o)^2 + (z_n - z_o)^2}
\end{equation}

Finally, the dynamic threshold for noise filtering is calculated (L.10).
\begin{equation}
  \label{eq:Hd}
  {\rm H}_d= {\rm H}_g \times {\rm C}_r \times ^{o}_n{\rm D}
\end{equation}
Here, ${\rm C}_r$ is the constant multiplication factor for range. If the average distance $\mu_n$ for each point ${\rm \bf p}_n$ is greater than the dynamic threshold ${\rm H}_d$, it is deemed noise and removed (L.12).
\end{itemize}

\subsection{Rough refinement with planes}
\subsubsection{Rough refinement of movement trajectory}
Using the plane point clouds ${\rm \bf \bar{P}}^p$ detected by the segmentation module, we correct the movement trajectory $\rm \bf S$ of the reference LiDAR $L_0$.
Specifically, we adjust the movement trajectory so that the normal $^{L_0}{\rm \bf n}_{t_0}$ and the distance $^G{\rm d}_{L_0,t_0}$ from the plane $^G{\rm \bf \bar{P}}^p_{L_0,t_0}$ detected by the reference LiDAR $L_0$ at time $t_0$ match the normal $^{L_0}{\rm \bf n}_{L_0,t_j}$ and the distance $^G{\rm d}_{L_0,t_j}$ from the plane $^G{\bf \bar{P}}^p_{L_0,t_j}$ detected at each timestamp $t_j$. Figure \ref{fig:refinement}(a) shows an image of the rough refinement of movement trajectory.
The change in pose ${\Delta {^{G}_{L_0}{\rm \bf T}}} = \{ {\rm \Delta ^{G}_{L_0}{\rm \bf R}} , {\rm \Delta ^{G}_{L_0}{\bf t}} \}$ is derived as follows:

\begin{equation}
  \label{eq:mt_delta_rot}
\Delta ^{G}_{L_0}{\rm \bf R} = {\rm M}({\rm acos}(^{L_0}{\rm \bf n}_{t_0} \cdot \ ^{G}_{L_0}{\rm \bf R}_{t_j} \ ^{L_0}{\rm \bf n}_{t_j})),
\end{equation}
\begin{equation}
  \label{eq:mt_delta_trans}
{\Delta ^{G}_{L_0}{\rm \bf t}} = {\rm \bf n}_{t_0}  (^Gd_{L_0,t_j} - {^Gd_{L_0,t_0}})
\end{equation}
where $\rm M()$ is a function that converts an angle into a rotation matrix.

\subsubsection{Rough refinement of extrinsic parameter}
\begin{figure}[!t]
\centering
\subfigure[Movement trajectory refinement]{
{\includegraphics[width=0.6\columnwidth, angle=0]{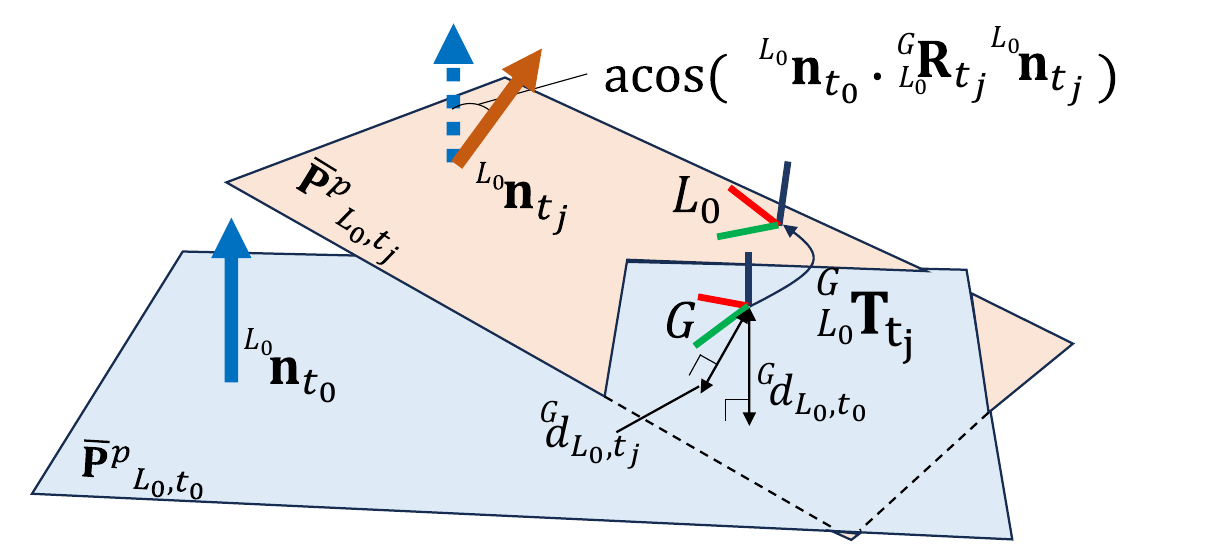}}}
\subfigure[Extrinsic parameter refinement]{
{\includegraphics[width=0.6\columnwidth, angle=0]{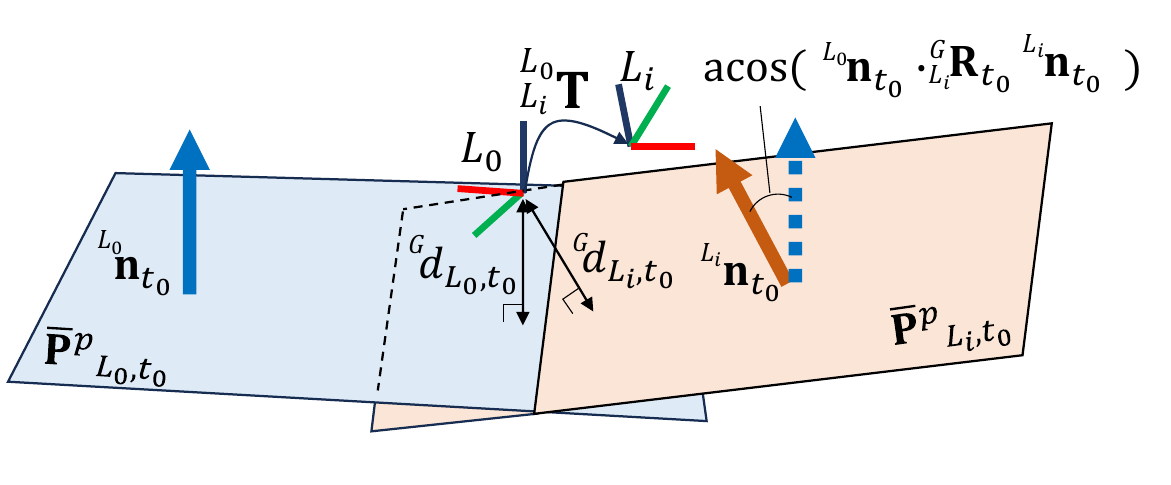}}}
\caption {Movement trajectory and extrinsic parameter refinement with planes}
\label{fig:refinement}
\end{figure}

After correcting the movement trajectory, the extrinsic parameters ${\rm \bf C} = \{^{L_0}_{L_1}{\rm \bf T}, \cdots, ^{L_0}_{L_{n-1}}{\rm \bf T} \}$ for each LiDAR are corrected similar conditions.
The correction of extrinsic parameters involves transforming the coordinates so that the normal ${\bf n}_{t_0}$ and distance $d_{t_0}$ to the plane at time $t_0$ for the reference LiDAR match the normal $^{L_i}{\bf n}_{t_0}$ and $^{L_i}d_{t_0}$ and distance $d_{t_0}$ observed by each LiDAR $\{L_i\}$.
Figure \ref{fig:refinement}(b) shows an image of the rough refinement of extrinsic parameter.
The change in pose $\Delta ^{L_0}_{L_i}\mathbf{T} = \{ \Delta ^{L_0}_{L_i}\mathbf{R} , \Delta ^{L_0}_{L_i}\mathbf{t} \}$ is derived as follows:

\begin{equation}
  \label{eq:ep_delta_rot}
{\Delta ^{L_0}_{L_i}{\rm \bf R}} = {M}(\rm {acos}(^{L_0}{\rm \bf n}_{L_0,t_0} \cdot ^G_{L_i}{\rm \bf R}_{L_i,t_0} \ ^{L_i}{\bf n}_{L_i,t_0})),
\end{equation}
\begin{equation}
  \label{eq:ep_delta_trans}
{\Delta ^{L_0}_{L_i}{\bf t}} = ^{L_0}{\rm \bf n}_{L_0,t_0}  (^Gd_{L_0,t_0} - ^Gd_{L_i,t_0})
\end{equation}

\subsection{Optimization with objects}
We refine the movement trajectory $\bf \hat{S}$ and the extrinsic parameters $\bf \hat{C}$ of each LiDAR using the object point clouds ${\rm \bf \bar{P}}^o$ detected by segmentation.
To solve the optimization problem in equation (\ref{eq:graph_optimization}), we define an error function.
Considering that the point clouds from LiDARs observing the travel surface may not capture objects with sufficient features, we have chosen to use a simple distance between point clouds without features as the error function.

In each optimization process, a kd-tree is constructed using the object point cloud $^G{\rm \bf \bar{P}}^o_{L_0}$ observed by the reference LiDAR, and the nearest neighbor $^{G}{{\rm \bf \hat{P}}^o_{L_0}}$ is found for the $^G{\rm \bf \bar{P}}^o_{L_i}$ transformed into global coordinates using the current movement trajectory $\bf \hat{S}$ and the extrinsic parameters $\bf \hat{C}$ of each LiDAR. The error is calculated as the sum of the distances between these point clouds.

\begin{equation}
\begin{split}
{\rm e}(^{G}\hat{\bf P}^o_{L_0},{^G{\bf \bar{P}}}^o_{L_i}) &= \sum^{N_o}_{n=1}{||^{G}{\rm \bf \hat{p}}_{n,L_0} - ^{G}{\rm \bf \bar{p}}_{n,L_i}||} \\
&= \sum^{N_o}_{n=1}{||^{G}{{\rm \bf \hat{p}}}_{n,L_0} - (^{G}_{L_i}{\rm \bf \hat{R}} \ {\rm \bf \bar{p}}_{n,L_i} + ^{G}_{L_i}{\rm \bf t})||},
\end{split}
  \label{eq:error_func}
\end{equation}
where $N_o$ is the number of points in the object point cloud.

Equation (\ref{eq:error_func}) is dependent on the poses $\bf \hat{S}$ and extrinsic parameters $\bf \hat{C}$ to be optimized.
By substituting these error functions into equation (\ref{eq:graph_optimization}) and recursively updating until convergence, similar to ICP, the optimized movement trajectory and extrinsic parameters are output. This paper solves the nonlinear optimization problem using the Ceres Solver \footnote{\url{http://ceres-solver.org/}} implemented in C++ with the Levenberg-Marquardt (LM) method.

\section{Experiments}
In this chapter, we evaluated the proposed methods using both simulation and real-world robot experiments.  We estimated the extrinsic parameters of two and four non-overlapping FoV LiDARs in a simulation environment.  We demonstrate that our proposed method, which includes rough refinement using the floor plane, can estimate extrinsic parameters more accurately compared to traditional methods.  Furthermore, we demonstrate that the proposed calibration method is applicable in the real world using data from an actual mobile robot.  We also show that the proposed noise filter module can effectively remove bleeding points noise, which occurs in real observations, compared to traditional methods.

\subsection{Simulation evaluation}
\begin{figure}[!b]
\centering
\subfigure[Multiple LiDAR]{
{\includegraphics[height=3.0cm, angle=0]{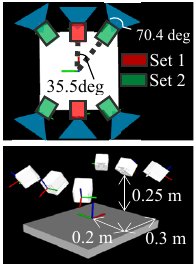}}}
\subfigure[Environment with objects]{
{\includegraphics[height=3.0cm, angle=0]{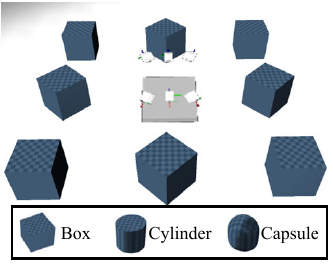}}}
\caption {Configuration of multiple LiDAR and environment in simulation. (a) The LiDARs are divided into two sets: Set 1 (red) with two LiDARs and Set 2 (green) with four LiDARs, each arranged to avoid overlap. (b) These LiDARs are moved in an environment where objects with a height of 0.25 m are placed around.
}
\label{fig:sim_settings}
\end{figure}

\subsubsection{Settings}
We created a mobile platform equipped with multiple LiDARs that have non-overlapping FoV in the simulation.  The configuration of the multiple LiDARs and simulation environement are shown in Fig. \ref{fig:sim_settings}. We used the Livox Mid-70 for the LiDARs\footnote{\url{https://www.livoxtech.com/mid-70}}.
We tested our proposed method using rotational motion in place and evaluated the error against the ground truth of the extrinsic parameters. The platform performed repeated stopping and rotating actions at poses where both the object and the floor surface could be observed, accumulating data at a total of 16 poses.

Verify that the proposed method is able to correctly adjust the extrinsic parameters even when randomly applied with a maximum noise of 0.2 m in translation and 5.0 deg in rotation.  Similarly, a maximum noise of 0.2 m and 3.0 deg is randomly added to the motion trajectory.  To demonstrate that estimation is possible with various objects, we used three types of objects: box, cylinder, and capsule.  In particular, cylinders and capsules are challenging objects for traditional feature-based methods due to their lack of planar features. The simulation environment was constructed using Mujoco\footnote{\url{https://mujoco.org/}}.

For accuracy comparison, we compared with feature-based method \cite{liu2022targetless} and the entire cloud-based method, which uses the entire point cloud as described in \cite{liu2021extrinsic} and methods that only use segmentation without rough refinement by planes (proposed w.o. RR). To maintain similar conditions, we used the same cost function as the proposed method for the entire cloud-based method, which is the sum of distances to points.  We evaluated the accuracy of the translation error and the rotation error.  Each method was optimized for 200 iterations during a evaluation process.

\subsubsection{Extrinsic error evaluation}
\begin{figure}[!t]
\centering
\subfigure[Set 1: two LiDARs]{
{\includegraphics[width=0.9\columnwidth, angle=0]{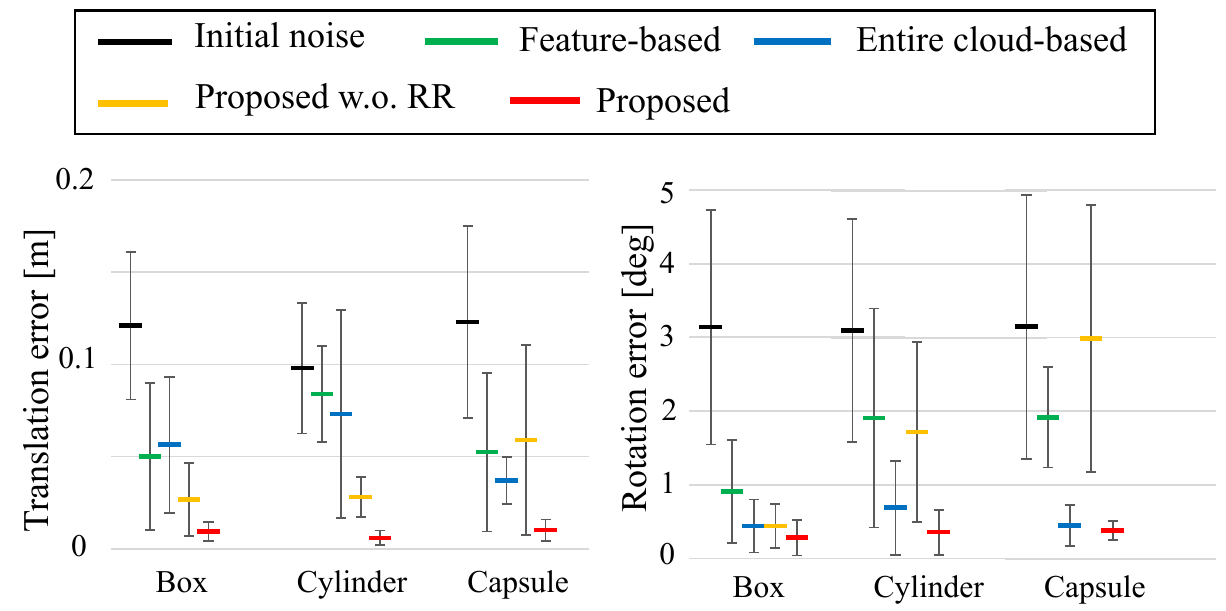}}}
\subfigure[Set 2: four LiDARs]{
{\includegraphics[width=0.9\columnwidth, angle=0]{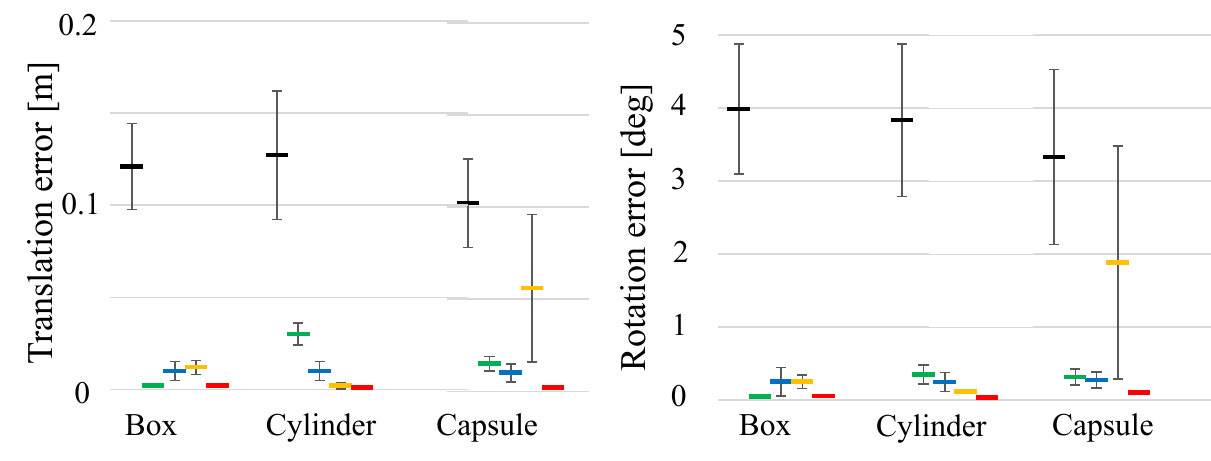}}}
\caption {
Comparison of translation and rotation error in each method.  The proposed method demonstrates the best accuracy regardless of the type of object.  In particular, it can be confirmed that the proposed method reduces the translation error to below 0.01 m and rotation error to below 0.3 deg in calibration using two LiDARs, while the accuracy of other methods decreases when using a cylinder or a capsule.  In particular, feature-based method tends to be less accurate when using cylinder or capsule compared to using boxes. This is discrepancy is likely due to the bias or lack of plane and edge features.  In set 2, where four LiDAR systems are employed, the proposed method also exhibited the highest accuracy. Although all the methods achieved relatively higher accuracy due to the abundance of data, methods without rough refinement sometimes converge to an incorrect position in the capsule. This further confirms the usefulness of the proposed rough refinement with planes.
}
\label{fig:sim_summary}
\end{figure}

Figure \ref{fig:sim_summary} shows the average values of translation error and rotation error for each method and object, which were tested 10 times. The evaluation results indicate that the proposed method achieves the highest accuracy compared to other methods, and could reduce the mean initial translation error of 0.12 m by up to 92\%, resulting in an translation error to below 0.01 m.

\begin{figure*}[!t]
\centering
\includegraphics[width=1.6\columnwidth, angle=0]{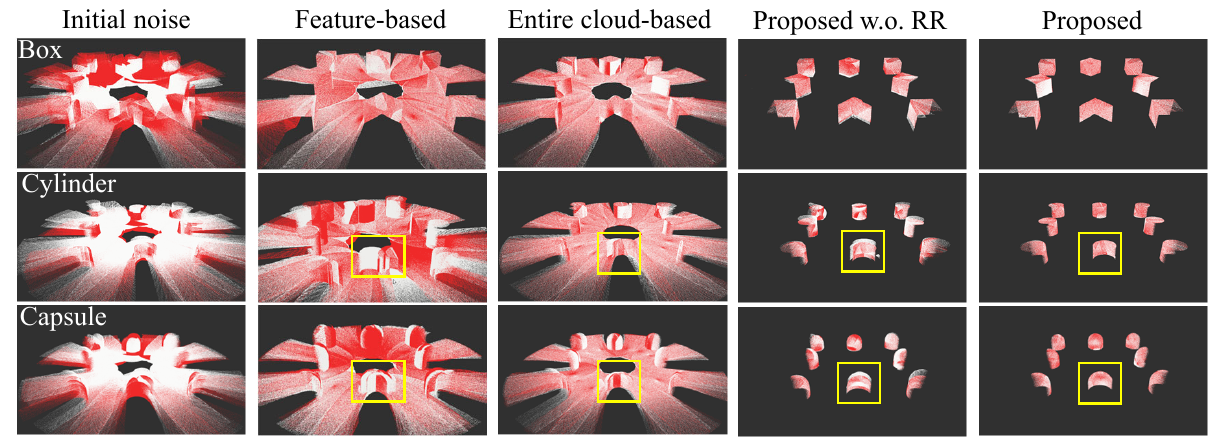}
\caption {
Comparison of multiple LiDAR point clouds before and after optimization in set 1. The white clouds represent reference LiDAR observations, and the red clouds represent calibration LiDAR observations. The conventional method that uses the entire cloud prioritizes matching the floor, resulting in a translation error when matching the object. The feature-based method has a larger error when using a cylinder or capsule than when using box. The proposed method without rough refinement converges to the wrong position. However, the proposed method accurately aligns two LiDAR point clouds regardless of the object type. The yellow squares indicate the areas where these differences are most pronounced.
}
\label{fig:sim_comparison_point_cloud}
\end{figure*}

Figure \ref{fig:sim_comparison_point_cloud} shows the point clouds before and after optimization in each method for Set 1.  It is shown that the proposed method can calibrate more accurately than conventional methods, regardless of object type, and that through rough refinement improves the calibration performance.

\subsubsection{
Comparison of iteration number to convergence
}
\begin{figure}[!h]
\centering
\includegraphics[width=0.75\columnwidth, angle=0]{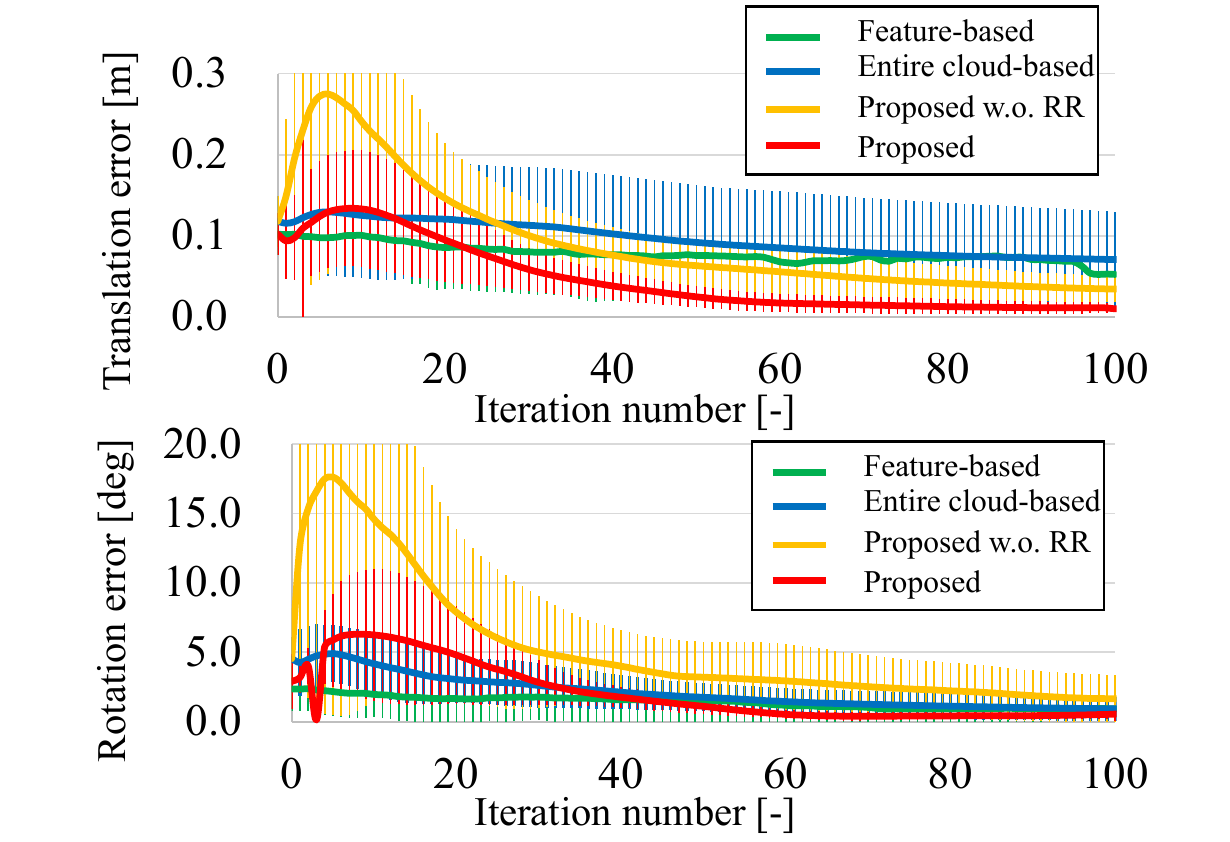}
\caption {
Comparison graph shows the iteration number and extrinsic parameter errors for each method.  The results of the graph indicate that rough refinement using planes leads to faster convergence.  It also demonstrates convergence in fewer iterations and higher accuracy compared to conventional methods that use the entire point cloud.
}
\label{fig:sim_comparison_iteration_num}
\end{figure}

 Figure \ref{fig:sim_comparison_iteration_num} shows a graph with iterations on the horizontal axis and translation error and rotation error on the vertical axis in set 1.  These results demonstrate that rough refinement by the floor plane in the proposed method reduces the number of iterations required for convergence and achieves the high accuracy.

\subsection{Real world evaluation}

\subsubsection{Settings}
We used our developed mobile robot Tachyon3\cite{tachyon3}, equipped with two Livox Mid-70s as shown in Fig. \ref{fig:proposed_image}.  By design, the front LiDAR was tilted at 35.4 deg and mounted at a height of 0.34 m from a base frame, while the rear LiDAR was tilted at 34.4 deg and also mounted at a height of 0.28 m from a base frame. The height of base frame was 0.54 m from a ground. Note that the design values were not reliably accurate and include errors that are difficult to measure, such as mounting errors.

As one of the evaluation, we defined the sum of distances between normalized point clouds as the metric error, since it was difficult to measure the true values of extrinsic parameters in real-world experiments.
\begin{equation}
\frac{\sum_{i=1}^{I}{\rm e(^{G}{\rm \bf \hat{P}}^{o}_{L_0},^{G}{\rm \bf \bar{P}}^{o}_{L_i})}}{\sum_{i=1}^{I}{(N_{L_i})}}
  \label{eq:metric_error}
\end{equation}
$N_{L_i}$ represent the number of point clouds observed by the LiDARs to be calibrated.

The robot was moved remotely to collect data, and extrinsic calibration was performed for each method.  Similar to the simulation, the initial position of the extrinsic parameters was randomly perturbed with a maximum noise of 0.2 m in translation and 5.0 deg in rotation from design value.  For the estimation of the motion trajectory, we used an in-house SLAM device mounted on the robot.
  We conducted optimization using the proposed method by utilizing both the SLAM trajectory and initial extrinsic parameters as input.
Figure \ref{fig:real_env}(a) shows the observation scenes. In the scene 1,  boxes that were 0.25 m height and these were placed aroud the robot. The various type of objects were placed around the robot in scene 2.

To evaluate the proposed noise removal module, we applied it to the object point clouds and calculated the precision, recall and F-score.
  To obtain ground truth for the box point cloud, we performed plane removal and manually cropped the point cloud based on the size of the box.
For comparison, we compared these values with traditional methods which are SOR\cite{SOR} and DSOR\cite{DSOR}.  We used the same parameters  $(\rm k, {\rm C}_s) = (20, 0.01)$ for all methods and used the same parameter ${\rm C}_r = 3.0$ for DSOR and proposed method.

\subsubsection{Noise filter module evaluation}
\begin{figure}[!h]
\centering
\includegraphics[width=0.50\columnwidth, angle=0]{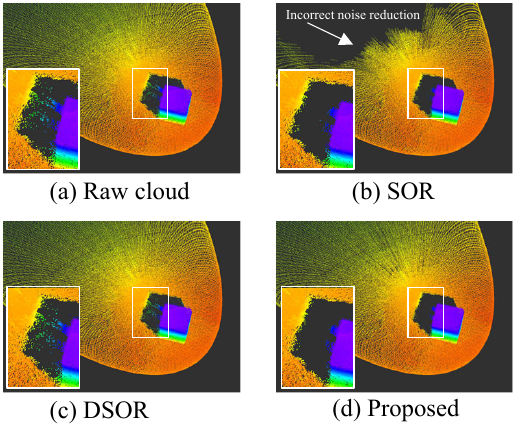}
\caption {
Comparison of noise in point clouds for each method. The raw cloud shows noise from the top of the box to the floor (indicated by the white square). SOR correctly removes noise but misidentifies sparse areas away from the sensor as noise (indicated by the white arrow). DSOR has fewer misidentification but does not detect enough noise to be removed. The proposed method removes the noise correctly with few misidentification.
}
\label{fig:comp_noise_filter}
\end{figure}

Table \ref{table:noise_filter_evaluation} summarizes the precision, recall and F-score for each method, and Figure \ref{fig:comp_noise_filter} shows the comparison results of the noise point clouds after applying each method in Scene 1.  The table shows that the proposed method had the highest F-score.  The SOR achieved high recall, but misrecognizes distant sparse floor surfaces, resulting in low precision. The DSOR reduced misrecognition of distant floor surfaces but had low recall because it did not fully remove dense noise at the center of the observations. It was confirmed that the proposed method correctly removes the high-density noise near the center of observation with few misidentification.

\subsubsection{3D map evaluation}
The Table \ref{table:real_metric_error} shows the average metric error before and after optimization.  It indicates that the proposed method reduces the average metric error. These results show that the two point clouds overlap with high precision and are integrated into one coordinate system. 

\begin{figure}[!t]
\centering
\includegraphics[width=0.60\columnwidth, angle=0]{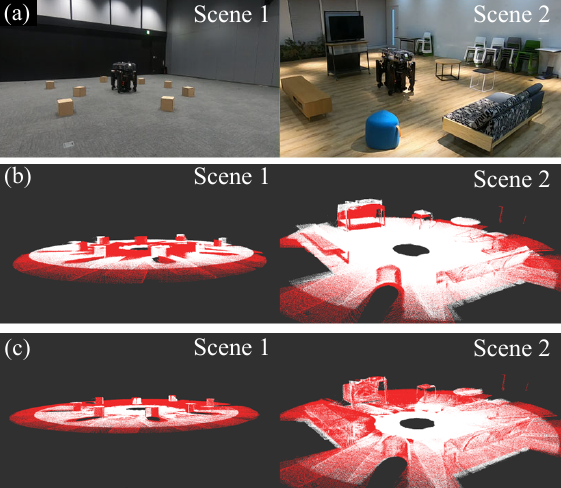}
\caption {
(a) Real world experiment environment. Comparison of (b) initial 3D map and (c) optimized 3D map after calibration. The white point clouds represent observations from the front LiDAR, which serves as the reference, while the red point clouds represent observations from the rear LiDAR.
}
\label{fig:real_env}
\end{figure}

Figures \ref{fig:real_env}(b) and \ref{fig:real_env}(c) show the point clouds before and after optimization for each scene.   The figures demonstrate that the proposed method can correctly integrate observations from two LiDARs with non-overlapping FoV in real world.

\begin{table}[!h]
\centering
\caption{
Comparison of precision and recall for each methods.
}
\begin{tabular}{ l | c c c}
\hline 
Methods  & Precision & Recall & F-score \\
\hline 
SOR         & 0.036    & 0.646  &0.068 \\
DSOR        & \bf 0.703    & 0.312  &0.433 \\
Proposed    & 0.638    & \bf 0.718  &\bf 0.676 \\
\hline 
\end{tabular}
\label{table:noise_filter_evaluation}
\end{table}

\begin{table}[!h]
\centering
\caption{Average metric error before and after optimization.}
\begin{tabular}{ l  c c}
\hline 
\multirow{2}{*}{Scene} &\multicolumn{2}{c}{Metric Error [m]} \\
\cline{2-3}
  & Initial & Optimized \\
\hline 
Scene 1 & 0.020    & 0.0010 \\
Scene 2 & 0.025    & 0.0008 \\
\hline 
\end{tabular}
\label{table:real_metric_error}
\end{table}

\section{Conclusion}
This paper proposes a calibration method for multiple LiDARs with non-overlapping fields of view. By leveraging floor information and point clouds obtained during operation, we achieve precise calibration of LiDARs directed towards the ground. Additionally, we incorporate a noise removal module that takes into account the scanning pattern, effectively reducing noise. Through simulation-based evaluations, we demonstrate the ability of our method to correct up to 92\% of randomly given noise when observing the floor and various objects using two or four LiDARs. Furthermore, using real hardware, we confirm the superior noise reduction effects of the noise removal module compared to conventional methods, contributing to the consistent creation of a 3D map. Our proposed method is cost-effective and does not require specialized calibration equipment, making it well-suited for future robot inspection facilities. However, it is important to note that our method assumes operation in flat environments and may experience decreased accuracy in irregular terrains such as outdoor settings. In our future research, we plan to focus on expanding the system to include multimodal devices integrated into robots.

\bibliographystyle{IEEEtran}
\bibliography{ref}

\begin{thebibliography}{10}
\providecommand{\url}[1]{#1}
\csname url@rmstyle\endcsname
\providecommand{\newblock}{\relax}
\providecommand{\bibinfo}[2]{#2}
\providecommand\BIBentrySTDinterwordspacing{\spaceskip=0pt\relax}
\providecommand\BIBentryALTinterwordstretchfactor{4}
\providecommand\BIBentryALTinterwordspacing{\spaceskip=\fontdimen2\font plus
\BIBentryALTinterwordstretchfactor\fontdimen3\font minus
  \fontdimen4\font\relax}
\providecommand\BIBforeignlanguage[2]{{%
\expandafter\ifx\csname l@#1\endcsname\relax
\typeout{** WARNING: IEEEtran.bst: No hyphenation pattern has been}%
\typeout{** loaded for the language `#1'. Using the pattern for}%
\typeout{** the default language instead.}%
\else
\language=\csname l@#1\endcsname
\fi
#2}}

\bibitem{liu2021low}
Z.~Liu, F.~Zhang, and X.~Hong, ``Low-cost retina-like robotic lidars based on
  incommensurable scanning,'' \emph{IEEE/ASME Transactions on Mechatronics},
  vol.~27, no.~1, pp. 58--68, Feb. 2021.

\bibitem{xue2019automatic}
B.~Xue, J.~Jiao, Y.~Zhu, L.~Zhen, D.~Han, M.~Liu, and R.~Fan, ``Automatic
  calibration of dual-lidars using two poles stickered with retro-reflective
  tape,'' in \emph{Proc. 2019 IEEE International Conference on Imaging Systems
  and Techniques (IST)}, Abu Dhabi, 2019, pp. 1--6.

\bibitem{9812062}
J.~Zhang, Q.~Lyu, G.~Peng, Z.~Wu, Q.~Yan, and D.~Wang, ``Lb-l2l-calib: Accurate
  and robust extrinsic calibration for multiple 3d lidars with long baseline
  and large viewpoint difference,'' in \emph{Proc. 2022 International
  Conference on Robotics and Automation (ICRA)}, Philadelphia, 2022, pp.
  926--932.

\bibitem{liu2021extrinsic}
X.~Liu and F.~Zhang, ``Extrinsic calibration of multiple lidars of small fov in
  targetless environments,'' \emph{IEEE Robotics and Automation Letters},
  vol.~6, no.~2, pp. 2036--2043, Apr. 2021.

\bibitem{liu2022targetless}
X.~Liu, C.~Yuan, and F.~Zhang, ``Targetless extrinsic calibration of multiple
  small fov lidars and cameras using adaptive voxelization,'' \emph{IEEE
  Transactions on Instrumentation and Measurement}, vol.~71, pp. 1--12, May
  2022.

\bibitem{yuan2021pixel}
C.~Yuan, X.~Liu, X.~Hong, and F.~Zhang, ``Pixel-level extrinsic self
  calibration of high resolution lidar and camera in targetless environments,''
  \emph{IEEE Robotics and Automation Letters}, vol.~6, no.~4, pp. 7517--7524,
  July 2021.

\bibitem{domhof2019extrinsic}
J.~Domhof, J.~F. Kooij, and D.~M. Gavrila, ``An extrinsic calibration tool for
  radar, camera and lidar,'' in \emph{Proc. 2019 International Conference on
  Robotics and Automation (ICRA)}, Montreal, 2019, pp. 8107--8113.

\bibitem{lin2020decentralized}
J.~Lin, X.~Liu, and F.~Zhang, ``A decentralized framework for simultaneous
  calibration, localization and mapping with multiple lidars,'' in \emph{Proc.
  2020 IEEE/RSJ International Conference on Intelligent Robots and Systems
  (IROS)}, NV, 2020, pp. 4870--4877.

\bibitem{heng2020automatic}
L.~Heng, ``Automatic targetless extrinsic calibration of multiple 3d lidars and
  radars,'' in \emph{Proc. 2020 IEEE/RSJ International Conference on
  Intelligent Robots and Systems (IROS)}, NV, 2020, pp. 10\,669--10\,675.

\bibitem{chang2023versatile}
D.~Chang, R.~Zhang, S.~Huang, M.~Hu, R.~Ding, and X.~Qin, ``Versatile
  multi-lidar accurate self-calibration system based on pose graph
  optimization,'' \emph{IEEE Robotics and Automation Letters}, vol.~8, no.~8,
  June 2023.

\bibitem{levinson2014unsupervised}
J.~Levinson and S.~Thrun, ``Unsupervised calibration for multi-beam lasers,''
  in \emph{Proc. Experimental Robotics: The 12th International Symposium on
  Experimental Robotics}, Berlin, 2014, pp. 179--193.

\bibitem{fernandez2015extrinsic}
E.~Fern{\'a}ndez-Moral, J.~Gonzalez-Jimenez, and V.~Ar{\'e}valo, ``Extrinsic
  calibration of 2d laser rangefinders from perpendicular plane observations,''
  \emph{The International Journal of Robotics Research}, vol.~34, no.~11, pp.
  1401--1417, May 2015.

\bibitem{jiao2019novel}
J.~Jiao, Q.~Liao, Y.~Zhu, T.~Liu, Y.~Yu, R.~Fan, L.~Wang, and M.~Liu, ``A novel
  dual-lidar calibration algorithm using planar surfaces,'' in \emph{Proc. 2019
  IEEE Intelligent Vehicles Symposium (IV)}, Paris, 2019, pp. 1499--1504.

\bibitem{jiao2019automatic}
J.~Jiao, Y.~Yu, Q.~Liao, H.~Ye, R.~Fan, and M.~Liu, ``Automatic calibration of
  multiple 3d lidars in urban environments,'' in \emph{Proc. 2019 IEEE/RSJ
  International Conference on Intelligent Robots and Systems (IROS)}, Macao,
  2019, pp. 15--20.

\bibitem{SOR}
R.~B. Rusu, ``Semantic 3d object maps for everyday manipulation in human living
  environments,'' \emph{KI-K{\"u}nstliche Intelligenz}, vol.~24, pp. 345--348,
  Aug. 2010.

\bibitem{ROR}
R.~B. Rusu and S.~Cousins, ``3d is here: Point cloud library (pcl),'' in
  \emph{Proc. 2011 IEEE International Conference on Robotics and Automation
  (ICRA)}, Shanghai, 2011, pp. 1--4.

\bibitem{DSOR}
X.~Zhong, R.~Lu, L.~Li, X.~Wang, and Y.~Zheng, ``Dsor: A traffic-differentiated
  secure opportunistic routing with game theoretic approach in manets,'' in
  \emph{Proc. 2019 IEEE Symposium on Computers and Communications (ISCC)},
  Barcelona, 2019, pp. 1--6.

\bibitem{DROR}
M.~H. Prio, S.~Patel, and G.~Koley, ``Implementation of dynamic radius outlier
  removal (dror) algorithm on lidar point cloud data with arbitrary white noise
  addition,'' in \emph{2022 IEEE 95th Vehicular Technology Conference:
  (VTC2022-Spring)}, 2022, pp. 1--7.

\bibitem{heinzler2020cnn}
R.~Heinzler, F.~Piewak, P.~Schindler, and W.~Stork, ``Cnn-based lidar point
  cloud de-noising in adverse weather,'' \emph{IEEE Robotics and Automation
  Letters}, vol.~5, no.~2, pp. 2514--2521, Dec. 2020.

\bibitem{seppanen20224denoisenet}
A.~Sepp{\"a}nen, R.~Ojala, and K.~Tammi, ``4denoisenet: Adverse weather
  denoising from adjacent point clouds,'' \emph{IEEE Robotics and Automation
  Letters}, vol.~8, no.~1, pp. 456--463, Sep. 2022.

\bibitem{besl1992method}
P.~J. Besl and N.~D. McKay, ``Method for registration of 3-d shapes,'' vol.
  1611, pp. 586--606, Apr. 1992.

\bibitem{chen1992object}
Y.~Chen and G.~Medioni, ``Object modelling by registration of multiple range
  images,'' \emph{Image and vision computing}, vol.~10, no.~3, pp. 145--155,
  Apr. 1992.

\bibitem{renyi1961measures}
A.~R{\'e}nyi, ``On measures of entropy and information,'' vol.~4, pp. 547--562,
  Jan. 1961.

\bibitem{tachyon3}
N.~Takasugi, M.~Kinoshita, Y.~Kamikawa, R.~Tsuzaki, A.~Sakamoto, T.~Kai, and
  Y.~Kawanami, ``Real-time perceptive motion control using control barrier
  functions with analytical smoothing for six-wheeled-telescopic-legged robot
  tachyon 3,'' 2023.

\end{thebibliography}

\end{document}